\documentclass{article}

\usepackage[final]{corl_2020} 

\usepackage{hyperref}       
\usepackage{url}            
\usepackage{booktabs}       
\usepackage{amsfonts}       
\usepackage{nicefrac}       
\usepackage{microtype}      
\usepackage{blindtext}
\usepackage{amsmath,amssymb}
\usepackage{gensymb}
\usepackage{algorithm}
\usepackage[algo2e]{algorithm2e}
\usepackage{color}
\usepackage{amsfonts}
\usepackage{graphicx}
\usepackage{tabulary,multirow,overpic,xcolor}
\usepackage{wrapfig}
\usepackage{sidecap}
\usepackage{wrapfig}
\usepackage{pifont}
\newcommand{\cmark}{\ding{51}}%
\newcommand{\xmark}{\ding{55}}%
\usepackage{xcolor,colortbl}
\usepackage{subcaption}
\usepackage{xspace}
\usepackage{authblk}

\newcommand{\app}{\raise.17ex\hbox{$\scriptstyle\sim$}}
\newcolumntype{x}[1]{>{\centering\arraybackslash}p{#1pt}}

\newlength\savewidth

\makeatletter\renewcommand\paragraph{\@startsection{paragraph}{4}{\z@}
  {.5em \@plus1ex \@minus.2ex}{-.5em}{\normalfont\normalsize\bfseries}}\makeatother

\newcommand{\abvdataset}{\mbox{TaskGrasp}\xspace}

\newcommand{\abvmethod}{\mbox{GCNGrasp}\xspace}
\newcommand{\abvdatasetcage}{\mbox{SG14000}\xspace}

\title{Same Object, Different Grasps: Data and Semantic Knowledge for Task-Oriented Grasping}

\author[1]{Adithyavairavan Murali}
\author[2]{Weiyu Liu}
\author[1]{Kenneth Marino}
\author[2,3]{Sonia Chernova}
\author[1,3]{Abhinav Gupta}

\affil[1]{\footnotesize The Robotics Institute, Carnegie Mellon University}
\affil[2]{\footnotesize Institute for Robotics and Intelligent Machines, Georgia Institute of Technology}
\affil[3]{\footnotesize Facebook AI Research}

\begin{document}
\maketitle


\begin{abstract}
Despite the enormous progress and generalization in robotic grasping in recent years, existing methods have yet to scale and generalize task-oriented grasping to the same extent. This is largely due to the scale of the datasets both in terms of the number of objects and tasks studied. We address these concerns with the \abvdataset dataset which is more diverse both in terms of objects and tasks, and an order of magnitude larger than previous datasets. The dataset contains $250K$ task-oriented grasps for $56$ tasks and $191$ objects along with their RGB-D information. We take advantage of this new breadth and diversity in the data and present the \abvmethod~framework which uses the semantic knowledge of objects and tasks encoded in a knowledge graph to generalize to new object instances, classes and even new tasks. Our framework shows a significant improvement of around $12\%$ on held-out settings compared to baseline methods which do not use semantics. We demonstrate that our dataset and model are applicable for the real world by executing task-oriented grasps on a real robot on unknown objects. Code, data and supplementary video could be found at \href{https://sites.google.com/view/taskgrasp/home}{https://sites.google.com/view/taskgrasp}.
\end{abstract}



\section{Introduction}
We have seen tremendous progress in the fundamental task of robotic grasping in recent years. State-of-the-art grasping algorithms have shown generalization to object instances \cite{mahler2017dex, qtopt2018, zeng2018robotic, pinto2016supersizing}, viewpoints \cite{levine2016learning}, DOF constraints \cite{6dofclutterICRA2020, 6dofgraspnet, ten2017grasp}, unknown environments \cite{robotsinhomenips2018} and even adversarial objects \cite{adversarialObjects2019}. The key reason for the success of these approaches is large-scale learning. Typically data is sampled from analytical approaches in simulation~\cite{mahler2017dex, 6dofgraspnet} or using a self-supervised framework~\cite{pinto2016supersizing,levine2016learning}. Despite these recent successes,
there is still a significant gap between how humans grasp objects and how robots perform picking. Most techniques plan for stable grasps assuming grasping to be the end goal. However, when humans grasp an object, we do so with a particular purpose in mind and grasping is just the first step as a means to that end. For example, when humans grasp a cup, we use the handle to drink from it though several other stable grasps exist. Humans also use objects creatively, 
such as scooping with a bowl or hammering with a heavy mug. Different tasks may require completely different grasps for the same object. To effectively operate in human homes and complete multiple tasks, a personal robot would have to learn from humans to generalize grasping to several tasks and skills beyond a tool's prototypical use. For instance, if the robot is cooking and needs to stir a pot of pasta but doesn't have a spoon at hand, it can use an alternate tool, such as a knife. To truly get to human-level grasping, we must study not just stable grasping or grasping for an object's primary use-case but rather how to grasp depending on both the task and the object.


What are the bottlenecks in Task-Oriented robotic grasping? The biggest hurdle is the need for human-labeled data. Unlike self-supervised or analytical approaches for which force sensing or contact models can provide labels for stable grasps, here we need humans to identify how an object can be grasped for multiple tasks. There has been a lot of recent work in this area, including~\citep{contactdb2019, liuICRA2020, fangRSS2018}. \citet{contactdb2019} used thermal imaging in a curated setup to study human grasping contacts on $50$ 3D printed objects for two tasks. \citet{fangRSS2018} proposed to jointly learn a task-oriented grasping network and manipulation policy in simulation with reinforcement learning and demonstrated the framework on two-goal tasks with two object categories. \citet{liuICRA2020} proposed a data-driven approach to learning the complex relationships between grasps, objects, tasks, and broadened semantic contexts. However, their approach required pixel-wise affordance segmentation \cite{do2018affordancenet} for a small set of known object categories, which is challenging to generalize and get supervision for. Despite this progress in learning from human grasping, there are still significant gaps, both from a data and methods perspective. On the data side, existing datasets are limited in terms of the number of object instances, but especially in the number of tasks and object classes collected. Yet, even if we scale the datasets, it is unclear if current approaches will generalize to new object categories and tasks in the real world. 
We tackle both problems: first, we collect a dataset that is diverse both in terms of objects and tasks and an order of magnitude larger than previous datasets.
Second, we exploit the semantic knowledge of objects and tasks to present a system that can generalize to new object instances, classes, and new tasks. To the best of our knowledge, this paper is one of the first efforts in demonstrating robust generalization in task-oriented grasping, especially with semantic knowledge.


More specifically, our first key contribution of this work is the collection of a large-scale dataset which we call \abvdataset. We increase the number of real objects from the current best of $50$ in prior works~\citep{contactdb2019} to $191$, and collect RGB-D point cloud observations and object-centric 6-DOF grasps for the task-oriented grasping problem. We also scale the number of object classes from $40$~\citep{contactdb2019} to $75$ and resolve each of these to the standard WordNet ontology~\citep{miller1995wordnet}. And perhaps most importantly, we scale the number of tasks from $2-7$ in prior works~\citep{liuICRA2020, contactdb2019, fangRSS2018} to $56$. This expanded dataset both gives a better benchmark for task-oriented grasping and allows us to study generalization by expanding the number of object categories and tasks.

In order to generalize to a new object or task, we need to have some prior semantics about it. For instance, if we knew that mugs and bowls were both containers, we might infer that we should apply the scoop action in a similar way. To this end, and for our second main contribution, we propose a method, called \abvmethod, that incorporates semantic knowledge into the end-to-end learning of task-oriented grasping from object point clouds. 
In particular, we use a Graph Convolutional Network (GCN)~\citep{kipf2017} to reason about a knowledge graph that encodes relations between objects and tasks, and further leverage word embeddings trained on large-scale language tasks to provide additional prior information. 
Our \abvmethod model 
shows a significant improvement of 12$\%$ and 3.5$\%$ on held-out tasks and object categories, respectively, compared to baselines which do not incorporate semantics. We also show that our method and dataset are applicable for actual robots by executing task-oriented stable grasps on a 7-DOF Sawyer Robot on unknown objects.

\vspace{-2mm}
\section{Related Work}
\vspace{-2mm}

{\bf Task-Oriented Grasping:} Prior work in Task-Oriented Grasping can be grouped into analytic methods, data-driven approaches using object state information, and frameworks learning from observations. Early work in analytic grasping proposed task wrench spaces with task-oriented grasp quality metrics \cite{borst2004grasp}. Data-driven approaches have been proposed to improve generalization, though a large body of work has relied on object state information. \citet{song2010learning} used generative Bayesian Networks to model the relations between objects, grasps and tasks; \citet{antanas2018semantic} and \citet{ardon2019learning} leveraged probabilistic logic languages to reason about grasp regions affording different tasks through semantic relations. However, both methods require grounding geometric information about objects to semantic representations and can only reason about semantic knowledge alone. A related line of work has used object parts and affordance detection \cite{do2018affordancenet, detry2017task, lakani2018exercising, kokic2017affordance}. \citet{do2018affordancenet} leveraged the affordances of object parts to define the correspondences between affordances and grasp types (e.g., rim grasp for parts with contain or scoop affordance).  \citet{detry2017task} trained a separate affordance detection model using synthetic data to detect suitable grasp regions for each task. While we do not provide explicit supervision for object affordance, we demonstrate that our model achieves an implicit understanding.

More recent works have learned task-oriented grasping from just RGB-D observations of objects. \citet{dang2012semantic} proposed an example-based approach which learns task-oriented grasps by storing visual and tactile data of grasps. \citet{hjelm2015learning} proposed a discriminative model based on visual features of objects. \citet{jang2017end} proposed an end-to-end learning method of grasping objects from specific categories in a bin. To accelerate learning from observations, there have been efforts in scaling datasets as discussed previously \cite{contactdb2019, liuICRA2020, fangRSS2018}. The computer vision community has also focused on annotating datasets for inferring human grasp pose estimation from visual data \cite{shan2020, huang2015, kokic2020} with the aim that it could be adapted to robotic grasping with kinematic retargeting. In this work, we propose an expanded dataset in terms of the number of object categories and tasks to study generalization. We also present a unified framework that jointly learns from semantic knowledge and geometric observations.

{\bf Semantic Knowledge in Vision:} The use of knowledge and knowledge graphs for visual reasoning has been well studied. Word embeddings from language has been used extensively ~\cite{frome2013devise}. Class hierarchies, such as WordNet~\citep{miller1995wordnet}, have often been used to aid in image recognition~\citep{zhueccv14}. More generally, knowledge graphs have found extensive use in visual classification and detection~\citep{marino17}, as well as zero-shot classification~\cite{Wang_zslCVPR2018}. We draw on many of the ideas from these works in Computer Vision, especially those related to word embeddings and graphs, and apply them to a robotics task and to 3D point cloud data.

{\bf Semantic Knowledge in Robotics:} In robotics, semantic knowledge has been used to help robots adapt to diverse and changing environments by providing abstractions that generalize across similar situations. Large-scale robotic knowledge bases, such as KnowRob~\citep{tenorth2017representations}, RoboBrain~\citep{saxena2014robobrain}, and RoboCSE~\citep{daruna2019robocse}, aimed to provide robots with extensive knowledge about objects, spaces, tasks, actions, and agents. Other methods leveraged more specific knowledge in a variety of robotic tasks, such as affordance learning~\citep{moldovan2012learning} and visual-semantic navigation~\citep{yang2018visual}. 
Similar to~\citet{antanas2018semantic} and~\citet{ardon2019learning}, we reason about semantic knowledge for task-oriented grasping, but we leverage semantic knowledge for generalization to novel object classes and tasks. 

\vspace{-2mm}
\section{Dataset}
\vspace{-2mm}


In this section we describe our dataset: \abvdataset, specifically its properties, collection and annotation methodology. 
As shown in Table \ref{tb:dataset}, \abvdataset is the largest and most diverse dataset for task-oriented grasping to date with respect to number of objects, categories and tasks.

\abvdataset contains $191$ individual household and kitchen objects comprising $75$ distinct object categories and varying in size, geometry, material, and visual appearance. Figure~\ref{fig:hierarchy} shows the class of each object and its proportion in the dataset. We collect RGB-D pointclouds for each object, and automatically annotate 250$K$ stable grasps. We also curate a list of $56$ everyday tasks that impose different semantic constraints on grasping and annotate for each grasp whether that grasp is appropriate for each particular task.


\begin{figure*}[t]
\centering
    \includegraphics[width=0.90\linewidth]{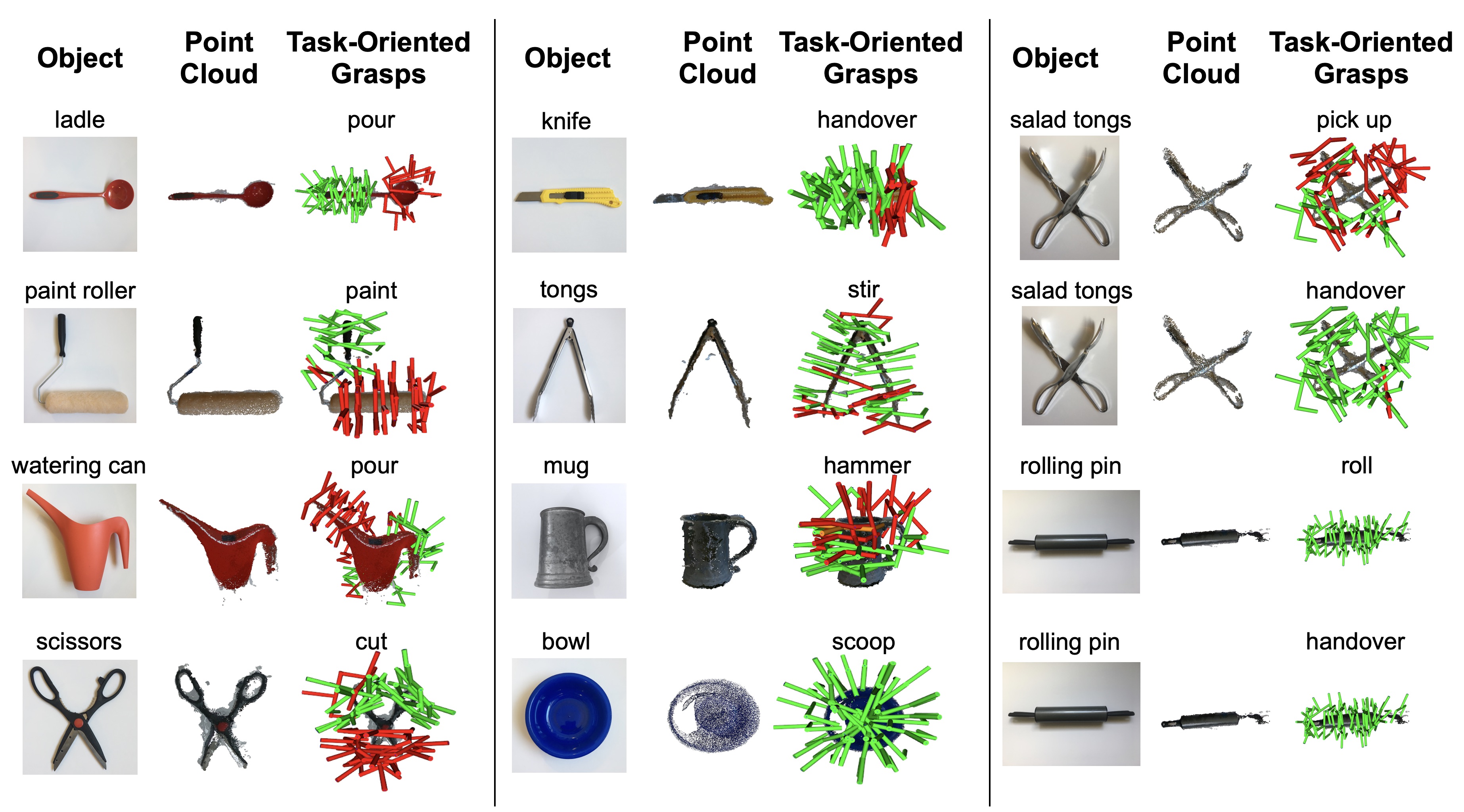}
    \vspace{-.2cm}
  \caption{Example point clouds and grasps from our \abvdataset dataset. Column 7-9 shows how grasps vary with tasks for a salad tongs (with higher diversity) and a rolling pin (with lower diversity). Green and Red means successful and incorrect task-oriented grasps respectively.}
  \vspace{-.6cm}
\label{fig:dataset}
\end{figure*}

\subsection{Data Acquisition on a Robot}
After selecting our $191$ objects by browsing various homegoods stores, we scan the objects to acquire their point clouds.
A Realsense D415 eye-in-hand camera mounted on a LoCoBot~\cite{pyrobot2019} is used for 3D scanning. The object is placed on a transparent mount in front of the robot, which is commanded to different poses along the object approach direction to capture point clouds from multiple viewpoints. This setup helps to capture more of the object geometry under self-occlusion, which in turn increases the coverage of grasp samples. The multi-view observations are registered using robot kinematics and further refined with the iterative closest point algorithm. After table plane segmentation, $600$ object-centric stable grasps are then sampled \cite{ten2018using} from the object point cloud. $25$ grasps are selected with farthest point sampling (to maximize grasp coverage) for annotation. These grasps are chosen as a representative, albeit limited, grasp set for the object to trade off between dataset size and budget.

\begin{table}[t]
\footnotesize
\centering
\caption{Comparing recent Task-Oriented Grasping Datasets}
\label{tb:dataset}
\begin{tabular}{c|cccc}
\hline
  & ContactDB \cite{contactdb2019}  & SG14000 \cite{liuICRA2020} & TOG-Net \cite{fangRSS2018} &  \abvdataset~(Ours)    \\
\hline
Semantic Knowledge                & \xmark       & \xmark & \xmark & \cmark \\
Object Categories & 40 & 5 & 2 & 75 \\
Objects & 50 & 44 & 18K (synthetic) & 191 \\
Tasks & 2 & 7 & 2 & 56 \\
Grasps & 3750 & 14K & 1.5M & 250K \\
Grasp Type & Contact Map & \textit{SE}(3) & Planar &  \textit{SE}(3) \\
\hline
\end{tabular}
\vspace{-.6cm}
\end{table}

\subsection{Data Annotation by Crowdsourcing}

We use Amazon Mechanical Turk (AMT) to crowdsource labels for the 250K stable grasps. Instead of exhaustively labelling each task-object combination ($\sim10K$) , we reduce the annotation cost with a two-stage procedure. We use the insight that the pre-condition for a task-oriented grasp is that the object has to be capable of the task in the first place.
First, we gather labels for whether a task is suitable for each object. Second, for this filtered subset of task-object combinations, we collect labels for the $25$ task-oriented grasps per object. To ensure annotation quality, we assign each labeling task to three annotators and use gold standard questions (questions that we know the answers to) to filter annotators with low accuracy. For both stages, we take a majority vote between the annotators.
We measure agreement with Randolph's free-marginal multirater kappa \cite{randolph2005free}. Kappa values for the two stages are $0.65$ and $0.62$ respectively ($0.0$ meaning agreement equal to chance, and $1.0$ indicating perfect agreement above chance), which suggests good agreement between annotators.

\subsection{Analysis}
In Figure~\ref{fig:dataset} we show prototypical examples from \abvdataset. We provide additional examples in the supplementary materials.

{\bf Diversity of Grasps:} As a result of the large number of objects and tasks, \abvdataset~contains a wide variety of task-oriented grasps. On average, each object is suitable for $7$ tasks. As shown in Figure~\ref{fig:dataset}, these tasks involve both prototypical (a ladle for pouring) and creative use of objects (tongs for stirring),  imposing drastically different semantic constraints on grasping. These examples also demonstrate the complex geometries presented in real world objects, which pose another challenge for generalization. 

We also quantitatively measure grasp diversity by analyzing the effect of tasks on grasps. Since different tasks provide different labels for the same set of stable grasps on each object, we compute Randolph's kappa \cite{randolph2005free} on these labels as a measure of agreement between tasks, i.e., how likely grasps for one task (e.g., stir) agree with grasps for another task (e.g., cut). Ranging from $0.19$ to $0.93$, kappa values of the objects suggest that the effect of tasks vary greatly for different objects. Column 7-9 in Figure~\ref{fig:dataset} show how grasps vary with tasks for a salad tongs with a kappa value of $0.38$ and a rolling pin with kappa value of $0.97$. In \abvdataset, $25\%$ of the objects have kappa values lower than $0.5$ and these objects require significantly different grasps for different tasks.

\begin{figure*}[t]
 \centering
     \includegraphics[width=0.55\linewidth]{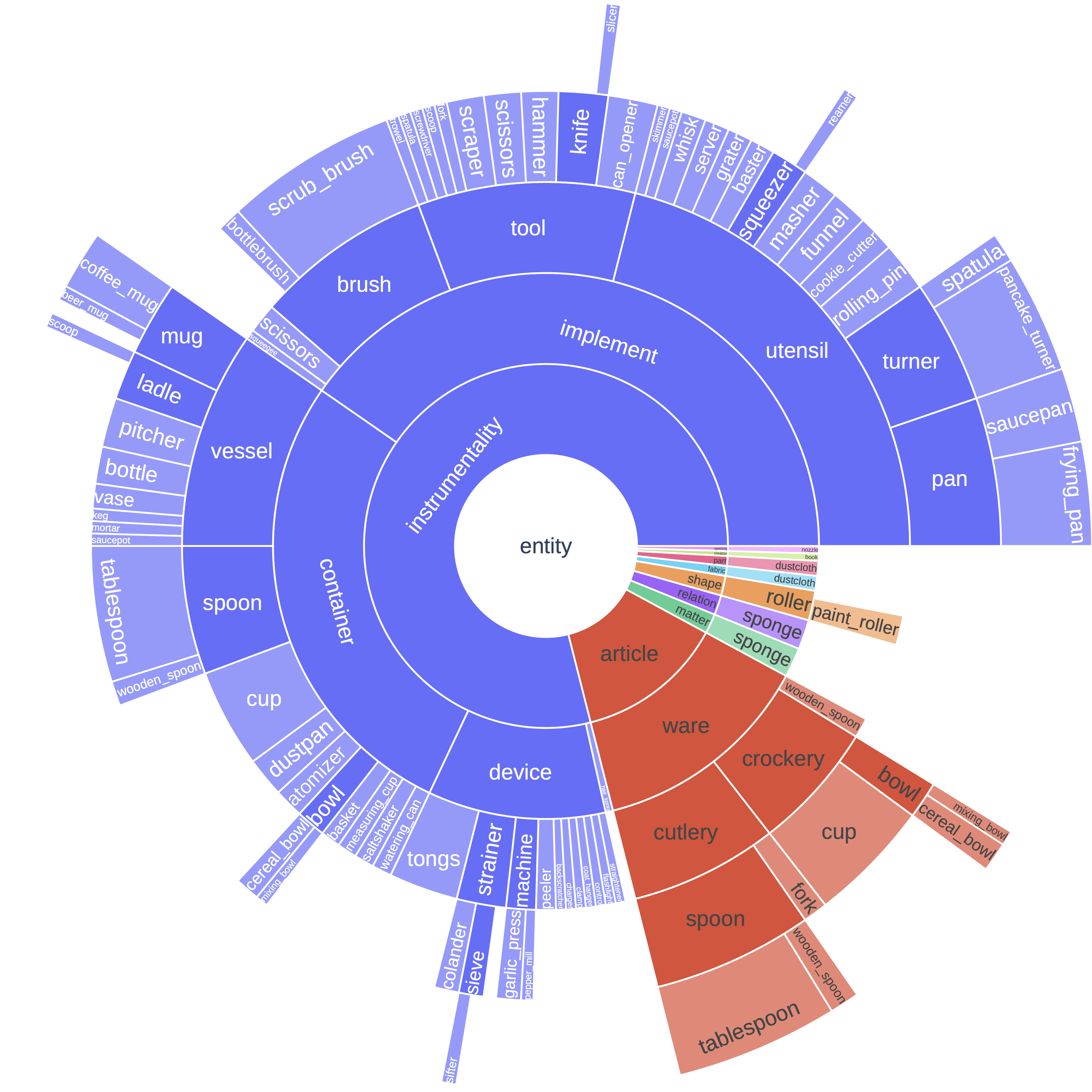}
   \caption{Semantic hierarchy of objects. Each level of the hierarchy is represented by one ring with the innermost circle as the root of the hierarchy. The angle of each segment is proportional to the number of objects. }
   \vspace{-.7cm}
 \label{fig:hierarchy}
 \end{figure*}

\begingroup


{\bf Semantic Knowledge of Objects and Tasks: } We also provide semantic knowledge about objects and tasks in the dataset. Objects are manually mapped to WordNet synsets \cite{miller1995wordnet} which represent a semantic hierarchy, as shown in Figure~\ref{fig:hierarchy}. Each of the 75 leaf synsets in the hierarchy represents a distinct object class and is linked to $2.5$ objects on average. Building on the hypernym paths from WordNet, the semantic hierarchy includes a rich set of object concepts interlinked by ``Is-A" relations. This provides useful semantic knowledge for task-oriented grasping as objects in the same subtree of the hierarchy often share similar functionalities or geometric properties. For example, mug, ladle, and bottle are in the vessel subtree and can all be used to hold liquid. In addition, we connect a task to an object class through ``Used-For" relations if any object in the class is considered suitable for the task from the first stage of our crowdsourcing. We provide a thorough breakdown of object counts, class hierarchies and used-for relations in the supplementary materials.

\endgroup
\vspace{-2mm}
\section{Task-Oriented Grasping with Semantic Knowledge}
\vspace{-2mm}

We consider the problem of generating grasps for task-oriented grasping given the object point cloud and task constraints. Specifically, we want to estimate the grasp distribution $P(G^{*}|X, \mathcal{T})$, where $X$ is the point cloud input, $\mathcal{T}$ are the  constraints imposed by goal tasks, and $G^{*}$ is the space of successful grasps. Following convention in related work \cite{ten2017grasp, 6dofgraspnet}, we represent grasps $g \in G^{*}$ as the grasp pose $(R, T) \in SE$(3) of a parallel-jaw gripper with its fingers open which when closing will lead to a stable grasp. We further factorize the estimation of $P(G^{*}|X, \mathcal{T})$ into 1) task-agnostic grasp sampling $P(G^{*}|X)$ and 2) task-oriented grasp evaluation $P(S|X, \mathcal{T}, g)$. The primary benefit of this factorization is that it allows us to leverage prior work in stable grasp generation. 

\begin{figure*}[t]
\centering
    \includegraphics[width=0.99\linewidth]{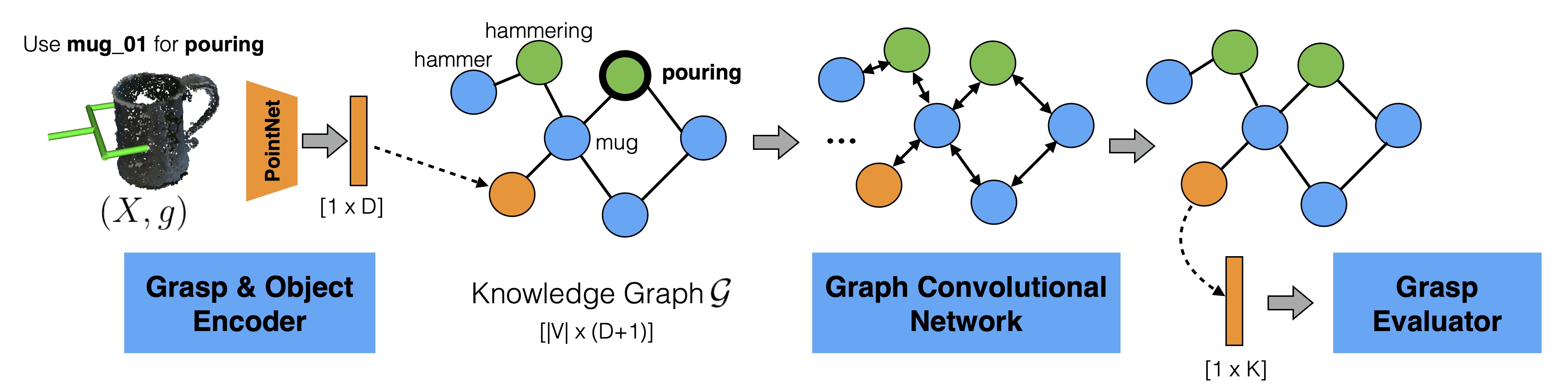}
  \caption{Overview of our Task-Oriented grasping framework using semantic knowledge graphs\vspace{-4mm}.}
\label{fig:system_overview}
\end{figure*}

In this section, we describe our method (\abvmethod)~for Task-Oriented grasping. Our method is composed of: (1) a Grasp and Object Encoder built on a PointNet++ architecture~\citep{pointnet} to encode the object point cloud and grasp, (2) a Graph Convolutional Network~\citep{kipf2017} which takes the encoded object shape and grasp as input as well as a knowledge graph $\mathcal{G}$ encoding the semantic relationships between object categories, tasks and hierarchies and (3) a Grasp Evaluator which outputs the final grasp prediction. See Fig ~\ref{fig:system_overview} for an overview of the framework.

\textbf{Grasp and Object Encoder:} Our input observations are object point clouds and we want to reason about $SE$(3) grasps. ~\citet{pointnet} proposed the PointNet++ architecture to efficiently represent 3D data which we use to learn a representation for the object point cloud and 6-DOF grasp poses. The grasp $g$ is defined in the object frame and six control points are selected on the gripper to form a gripper point cloud $X_{g}$. Similar to~\citet{6dofgraspnet}, $X_{g}$ is concatenated with the object point cloud $X$ with an extra latent indicator vector to represent whether a point is part of the gripper or the object. The PointNet layer reasons about the relative spatial information between the grasp and the object. It outputs a embedding which is used initialize the grasp node (orange in Fig \ref{fig:system_overview}) in the graph.


\textbf{Graph Convolutional Network:} We use the standard Graph Convolutional Network (GCN) model from \citet{kipf2017}, which is a neural network structured on the shape of the input graph. By structuring a neural network to pass information between adjacent nodes, we use the input graph to correctly reason about the relationship between the object classes and the target task. The first input of a GCN is the graph itself $\mathcal{G} = (V, E)$. In our application, we use a knowledge graph constructed from two sources: the task-object class relationships in our dataset and the object hierarchy from WordNet \cite{miller1995wordnet}. The grasp nodes (orange in Fig \ref{fig:system_overview}) are added online to the existing knowledge graph $\mathcal{G}$ by connecting edges to the corresponding object class nodes. The graph is represented as a binary adjacency matrix $A$, which we normalize to obtain $\hat{A}$ following~\cite{kipf2017}. Each node of the graph also has a $D-$dimensional embedding which is used by the GCN. The target tasks are specified using an extra indicator latent variable that is concatenated with this embedding to get a vector of size $D+1$. Except for the grasp nodes, we initialize the matrix with the word embeddings corresponding to each concept in the knowledge graph (e.g. ``mug''). We use ConceptNet numberbatch ~\cite{speer2017conceptnet} for the word embeddings. The embedding vectors are stacked across nodes along the first dimension to get the input matrix $\mathcal{X} \in \mathcal{R}^{|V| \times (D+1)}$.  The output of the GCN are $K$-dimensional embeddings for each node $\mathcal{Z} \in \mathcal{R}^{|V| \times K}$. The node embeddings are propagated to their neighbours using message passing in each convolutional layer:

\vspace{-.2cm}
\begin{equation}
    H^{(l+1)} = \sigma (\hat{A} H^{(l)} W^{(l)})
\label{eq:gcn}
\end{equation}
\vspace{-.3cm}

where $\sigma$ is the ReLU activation function, $H^{(0)} = \mathcal{X}$ and $H^{(L)} = \mathcal{Z}$ where $L$ is the number of layers.

 
\textbf{Grasp Evaluator:} After the GCN, we are left with a node-level embedding $\mathcal{Z}$. We use the embedding corresponding to the grasp node $z_{g}$ to train the final grasp evaluator $P(S|z_{g})$, where $S$ is the grasp score. This module has three fully connected layers with $K$ units and a final sigmoid layer. The entire model, including the grasp and object encoder, GCN and grasp evaluator, is optimized with ADAM using a binary cross entropy loss.


\textbf{Implementation Details:} 
The point clouds were downsampled to $4096$ points during training. They were also mean centered and unit-scaled. The PointNet module consists of three set abstraction layers and the number of points sampled are 512, 128 and all points. The set abstraction layers are followed by three fully connected layers with sizes [1024, 512, $D$]. Each set abstraction layer has three fully connected layers to learn features. The point clouds were perturbed with random rotations, jitter and dropout for data augmentation and to build robustness when testing on novel objects in unknown poses. We choose $D$=$300$ and $K$=$128$, and $L$=$6$ as the parameters for our GCN network.
\vspace{-2mm}
\section{Experimental Evaluation}
\vspace{-2mm}

%
\subsection{Zero-Shot Generalization}
A central goal of both our dataset and our method is to show that we can learn task-oriented grasping models which generalize to novel objects, classes and tasks. In an ideal robotics system, we should be able to correctly grasp a novel object from a novel object class, or even grasp for a novel task. To test this, we measure our system and baselines in three different held-out test settings: held-out object instances, held-out object categories, and held-out tasks.

These held-out settings are of increasing difficulty in terms of zero-shot generalization. For each setting, we perform $k$-fold cross validation ($k$=4), such that each category (a task, object class, or object instance, based on the setting) will be held out exactly once. In each fold, grasps from $25\%$ of the categories will be used for testing while remaining grasps will be used for training and validation. 

In all experiments, we only evaluate tasks that are valid for a given input object class. This makes sense from an evaluation perspective as it separates the problem of predicting applicable tasks for objects from task-driven grasping. It also makes the comparison to methods using object-task information fair since the models do not have to decide whether the object-task pair is valid.


\textbf{Evaluation Metrics:} 
Since $k$-fold cross validation in any held-out setting will evaluate all grasps in the dataset, we can compute Average Precision (AP) scores for any category, i.e., any object instance, object class, or task. We then compute an mAP averaged over object instances, mAP averaged over object classes, and mAP over tasks. We show all three metrics for each of our three settings in Tables~\ref{tb:instance_generalization},\ref{tb:class_generalization},\ref{tb:task_generalization}, but emphasize the mAP metric that corresponds to what category is being held out.


\textbf{Baselines:} 
We compare our approach to the following models: (1) Random, which represents grasping strategies that focus on grasp stability and ignore task constraints. Results are averaged over five random seeds. (2) Semantic Grasp Network (SGN), which learns to reason about context of grasps (e.g., constraints imposed by objects and tasks) from data. This model is adapted from \cite{liuICRA2020}, with the difference that the input to the model is replaced with geometric embedding from our shape encoder and word embeddings of the task and the object class. Note that embeddings of tasks and object classes are both learned from training data. (3) SGN + \textit{word embedding}, which uses ConceptNet \cite{speer2017conceptnet} numberbatch as pretrained word embeddings for object classes and tasks. 

\begin{table*}[t]
\caption{Results on \abvdataset}
\vspace{-.2cm}
\begin{subtable}{.45\linewidth}\centering\scriptsize
    \caption{Object Instance Generalization
    \vspace{-.2cm}
    \label{tb:instance_generalization}}
    \begin{tabular}{cccc}
    \hline
    \multirow{2}{*}{\textbf{Model}} & 
    \multicolumn{3}{c}{\textbf{Test Performance (mAP)}} \\
    \cmidrule(lr){2-4}
    & \cellcolor{gray!25} Instances & Classes & Tasks    \\
    \hline
    Random & \cellcolor{gray!25}59.75 & 60.28 & 54.76 \\
    SGN \cite{liuICRA2020} & \cellcolor{gray!25}78.51 & 75.08 & 68.8\\
    SGN + word embedding & \cellcolor{gray!25}79.74 & 77.91 & \textbf{74.36}\\
    \abvmethod~(ours) & \cellcolor{gray!25} \textbf{80.25} & \textbf{77.94} & 73.71\\
    \hline
    \end{tabular}
\end{subtable}
\begin{subtable}{.45\linewidth}\centering\scriptsize
    \caption{Object Class Generalization \label{tb:class_generalization}}
    \vspace{-.2cm}
    \begin{tabular}{cccc}
    \hline
    \multirow{2}{*}{\textbf{Model}} & 
    \multicolumn{3}{c}{\textbf{Test Performance (mAP)}} \\
    \cmidrule(lr){2-4}
    & Instances & \cellcolor{gray!25}Classes & Tasks    \\
    \hline
    Random & 59.32 & \cellcolor{gray!25}58.73 & 52.27 \\
    SGN \cite{liuICRA2020} & 74.2 & \cellcolor{gray!25}72.95 & 62.55 \\
    SGN + word embedding & 77.21 & \cellcolor{gray!25}75.51 & \textbf{63.73} \\
    \abvmethod~(ours) & \textbf{78.81} & \cellcolor{gray!25}\textbf{76.57} & 57.36 \\
    \hline
    \end{tabular}
\end{subtable}
\begin{subtable}{\linewidth}\centering\scriptsize

    \vspace{.2cm}
    \caption{Task Generalization \label{tb:task_generalization}}
    \begin{tabular}{cccc}
    \hline
    \multirow{2}{*}{\textbf{Model}} & 
    \multicolumn{3}{c}{\textbf{Test Performance (mAP)}} \\
    \cmidrule(lr){2-4}
    & Instances & Classes & \cellcolor{gray!25}Tasks    \\
    \hline
    Random & 59.06 & 58.24 & \cellcolor{gray!25}52.37 \\
    SGN \cite{liuICRA2020} & 75.17 & 71.59 & \cellcolor{gray!25}63.35 \\
    SGN + word embedding & 78.06 & 74.49 & \cellcolor{gray!25}70.55 \\
    \abvmethod~(ours) & \textbf{81.5} & \textbf{79.56} & \cellcolor{gray!25}\textbf{76.01} \\
    \hline
    \end{tabular}
\end{subtable}
\vspace{-.2cm}
\label{tb:allresults}
\end{table*}


\begin{table}[t]
\scriptsize
\centering
\caption{Ablation on Semantic Knowledge}
\label{tb:kbablation2}
\begin{tabular}{cccccc}
\hline
  \multirow{2}{*}{\textbf{Model}} & 
  \multicolumn{2}{c}{\textbf{Graph}} &
  \multicolumn{3}{c}{\textbf{Held-out Setting}} \\
  \cmidrule(lr){2-3}\cmidrule(lr){4-6}
  & Nodes      & Edges   & Task & Class & Instance \\
\hline
GCN + tasks + WordNet & 345 & 989 & 76.01 & \bf{76.57} & 80.25  \\
GCN + tasks  & 131 & 693 & \bf{77.54} & 75.86 & \bf{81.46} \\
GCN + WordNet  & 155 & 106 & 71.77 & 70 & 78.66 \\
\hline
\end{tabular}
\vspace{-.6cm}
\end{table}

\vspace{-2mm}
\subsection{Analysis}
\vspace{-2mm}
First, to get context for our results in Table \ref{tb:allresults}, we see that random grasp prediction achieves approximately 50-60\% accuracy, establishing a floor for the other methods. Because the number of positive and negative grasps in the dataset is about even, random guessing is able to achieve a seemingly high mAP. In a dataset with more negatives we would expect this number to be much lower.

Our method outperforms baselines in all three settings. This confirms that our method can effectively leverage the knowledge graph to generalize 
to novel object instances, object classes, and tasks. SGN + \textit{word embedding} also outperforms SGN, suggesting that implicit distributional knowledge provides a prior that is useful for generalization. Despite the benefit of distributional knowledge, it still only represents semantic similarities between concepts. In contrast, the knowledge graph directly stores relations between the relevant objects and tasks, and exploiting this additional structured knowledge allows our model to achieve better zero-shot generalization than SGN + \textit{word embedding}. 

When comparing our method with SGN and SGN + \textit{word embedding}, we observe increasingly larger margins in performance from the held-out instance to the held-out class setting. As objects from different classes have more variance in terms of geometric and visual features than objects from the same class, semantic knowledge becomes more important in unifying these objects. The difference in performance between our method and these two baselines on the held-out task setting reached 12.6$\%$ and 5.46$\%$ respectively, affirming that semantic knowledge is especially crucial for generalizing disparate constraints from different tasks. 

\textbf{Ablations on Knowledge Graph:} 
%
We investigated how performance is affected by changing the knowledge graph used in our model. Specifically, we compared the default knowledge graph with a knowledge graph containing only the semantic hierarchy of objects and a knowledge graph containing only the relations between object classes and tasks. The results from the three held-out settings are summarized in Table \ref{tb:kbablation2} (we only show the mAP metrics corresponding to the held-out category). From these results, we observe that edges between object classes and tasks were the most important knowledge for generalizing to novel tasks and instances, though every task we tested was valid for the target object class. This suggests that knowledge about which objects could generally be used for which tasks provide important information for discovering similarities between tasks. In the held-out object class setting, additional knowledge from the object hierarchy helped generalize to novel object classes by associating known classes and novel classes through the WordNet hierarchy.

\begin{figure*}
\centering
    \includegraphics[width=0.99\linewidth]{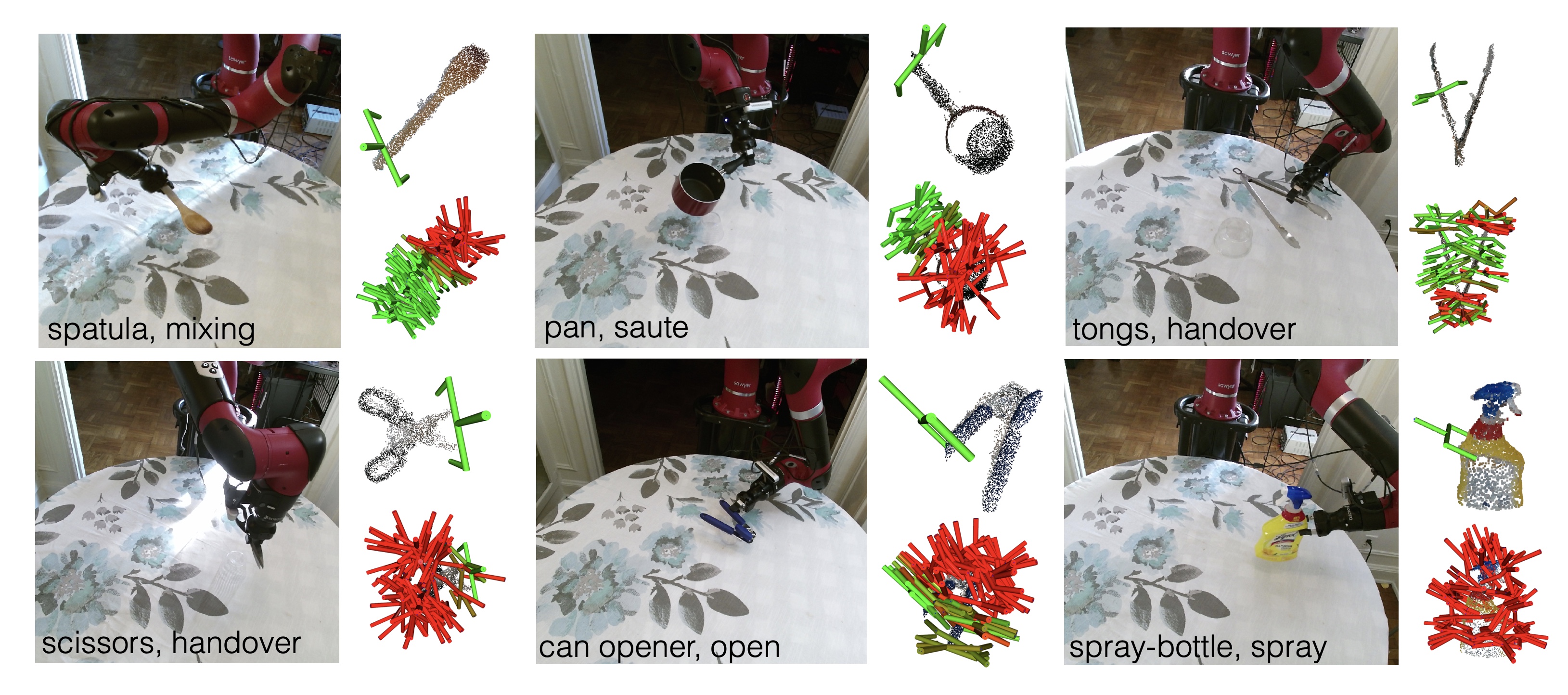}
    \vspace{-.4cm}
  \caption{Robot executions of example task-oriented grasps on unknown objects. For each execution, the top 3D visualization shows the grasp that was executed (which had the best evaluator score) and the bottom shows all the stable grasp candidates colored by their scores (green is higher).}
  \vspace{-.4cm}
\label{fig:robot}
\end{figure*}

\vspace{-2mm}
\subsection{Real Robot Evaluation}
\vspace{-2mm}

We run experiments to show that 
our approach and dataset transfer to a real robot. We test our approach on novel objects not from the dataset and in unknown poses. We place each object (without clutter) on a table in front of the robot. After table plane segmentation to obtain the object point cloud, $600$ stable grasps are sampled and $50$ candidates are selected using farthest point sampling for evaluation. We evaluate the grasps on our best performing \abvmethod model from the held-out task ablations (Table \ref{tb:allresults}). Our hardware setup comprises of a 7-DOF Sawyer Robot with a 2-fingered Robotiq gripper and a Intel Realsense D415 RGB-D camera mounted on the gripper wrist. Inference for the $50$ grasps takes around $3s$ on a desktop with an NVIDIA GTX 1080 Ti GPU and the grasp with the best score is executed on the robot. Fig \ref{fig:robot} shows the executed task-oriented grasps on unknown objects. Even though our dataset objects were collected only in one canonical pose, our approach is able to generalize to new grasps and in unknown poses due to data augmentation during training. Based on the grasp evaluator scores from Fig \ref{fig:robot}, our model is also able to interpolate between modes in the continuous $SE$(3) space to reason about task-oriented grasping. One failure mode of our work is that it does not generalize to categories (like the spray bottle in Fig \ref{fig:robot} in the bottom right) with limited training data. A future work is to balance the dataset in terms of object categories.

\vspace{-2mm}
\section{Conclusion}
\vspace{-2mm}

We present the \abvdataset dataset to study generalization in Task-Oriented grasping. The dataset is diverse and an order of magnitude larger than previous datasets. We also present a framework for jointly learning from geometric observations and semantic knowledge to generalize to new object instances, classes and even new tasks. Future work could explore recent techniques in automatic knowledge graph generation \cite{comet2019} for grasping tasks. While we collected real point cloud data of objects, we could convert the point clouds to meshes or acquire shape models from large online repositories to use in physics simulators. This could expand the scope of the dataset for sim2real transfer and to even learn task policies in simulation conditioned on the task-oriented grasps like in prior work \cite{fangRSS2018}.






\small
\bibliography{references}  

\appendix

\section{Appendix}






We provide additional analysis, dataset examples and experimental details in the appendix. We first demonstrate that models trained on our \abvdataset dataset transfer to other task-oriented grasping datasets, namely \abvdatasetcage proposed by \citet{liuICRA2020}. Second, we show more analysis on our dataset and model predictions on different tasks. Lastly, we describe our procedure and graphical interface for annotating task-oriented grasps on Amazon Mechanical Turk (AMT).

\subsection{Comparison to \abvdatasetcage}
We want to demonstrate that grasping models trained on our \abvmethod dataset generalize to other task-oriented grasping datasets. We show transfer learning results on \abvdatasetcage, since it has the most similar setting by providing objects with their corresponding point clouds and grasps in $SE$(3). Since \abvdatasetcage does not come with any semantic knowledge, we use the Semantic Grasping Network (SGN) + \textit{word embedding} as the backbone model instead of \abvmethod. \abvdatasetcage is significantly smaller and less diverse with 14K grasps. The five object categories and seven tasks were resolved to WordNet synsets to have complete overlap with \abvdataset. The test dataset was held-out based on grasps, hence may include known object classes and tasks during evaluation. The model trained on \abvdatasetcage performed well when tested on itself. However, it failed to generalize to the more diverse \abvdataset with only a 17\% increase over a random baseline. It is not surprising that the model trained on \abvdataset was able to generalize to the held-out test set in \abvdataset. It also performed well when tested on the \abvdatasetcage test set though it did not outperform the model trained on \abvdatasetcage. This is owing to several reasons. First, the point clouds in \abvdatasetcage were incomplete with a lot of self-occlusions (since objects were scanned from just a single view) whereas our point clouds are constructed based on scans from multiple view points. This could affect the performance of the Object and Grasp Encoder based on PointNet \cite{pointnet}. Second, \abvdatasetcage has a dataset bias since it models the effects of material and object state on grasps, while we focus on object geometry. Another reason could be dataset imbalance in \abvdataset as we do not have sufficient quantities of certain categories (bowls, bottles) in comparison. Lastly, \abvdatasetcage has some grasps in free space (which we filtered out in our dataset) where our model predicts a high score. This can be corrected by adding unstable grasps as hard negatives during training, similar to prior work \cite{6dofgraspnet, 6dofclutterICRA2020}.

\begin{table}[H]
\footnotesize
\centering
\caption{Cross generalization on \abvdataset and \abvdatasetcage}
\label{fig:sg14k}
\begin{tabular}{c|cc}
\hline
 \multirow{2}{*}{\textbf{Train Dataset}} & \multicolumn{2}{c}{\textbf{Held-out Test Grasps (mAP)}} \\
  & \abvdataset      & SG14000 \cite{liuICRA2020}  \\
\hline
\abvdataset & 76.2       & 52.3     \\
SG14000     & 25.1       & 62.7     \\
Random & 7.9 & 24.8 \\
\hline
\vspace{-4mm}
\end{tabular}
\end{table}

\begin{figure*}[t]
\centering
    \includegraphics[width=0.95\linewidth]{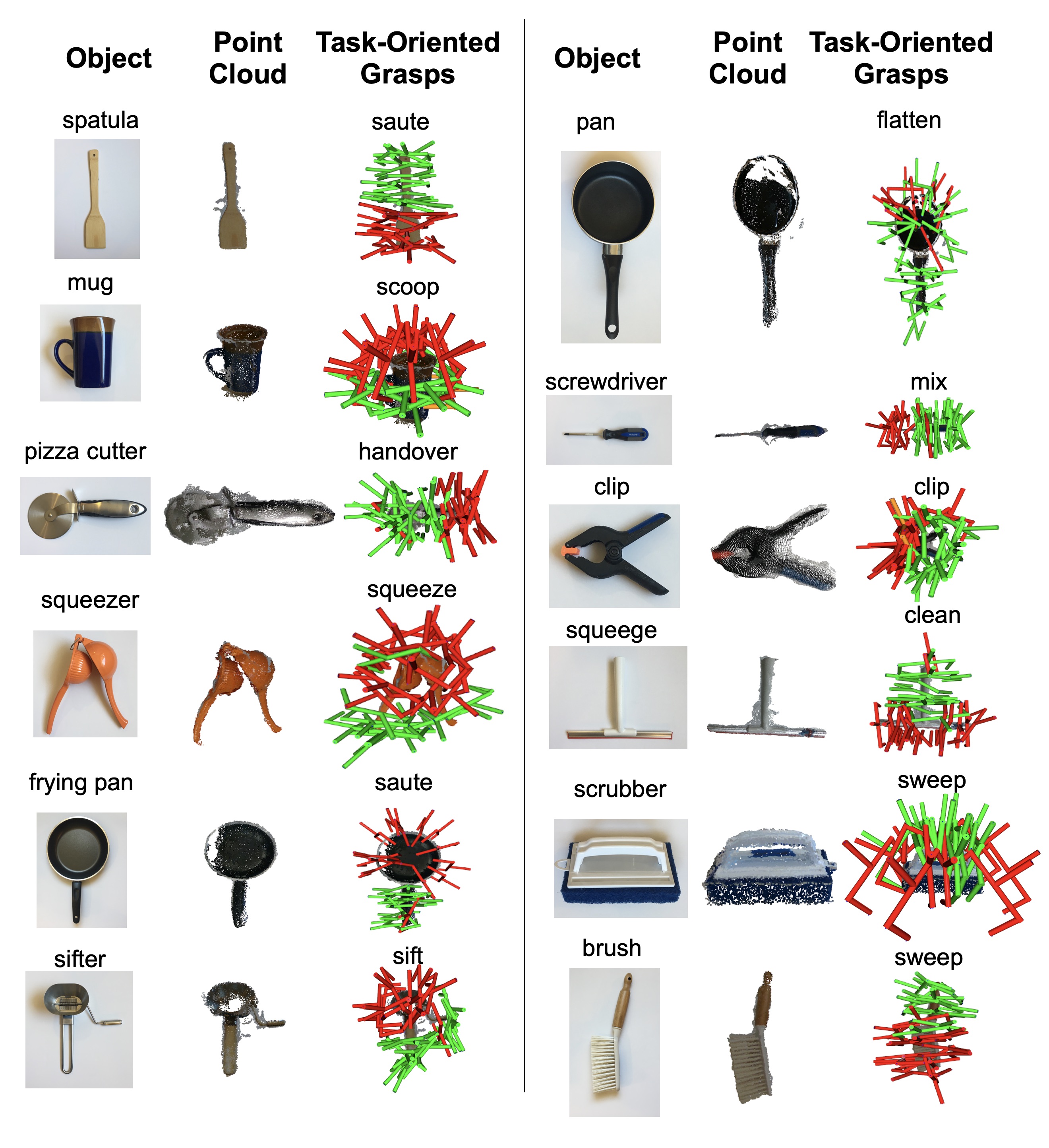}
    \vspace{-.2cm}
  \caption{Example point clouds and grasps from our \abvdataset dataset. Grasps colored in green and red are successful and incorrect respectively for task-oriented grasping.}
  \vspace{-.6cm}
\label{fig:extra_dataset}
\end{figure*}

\begin{figure}
\centering
    \includegraphics[width=0.85\linewidth]{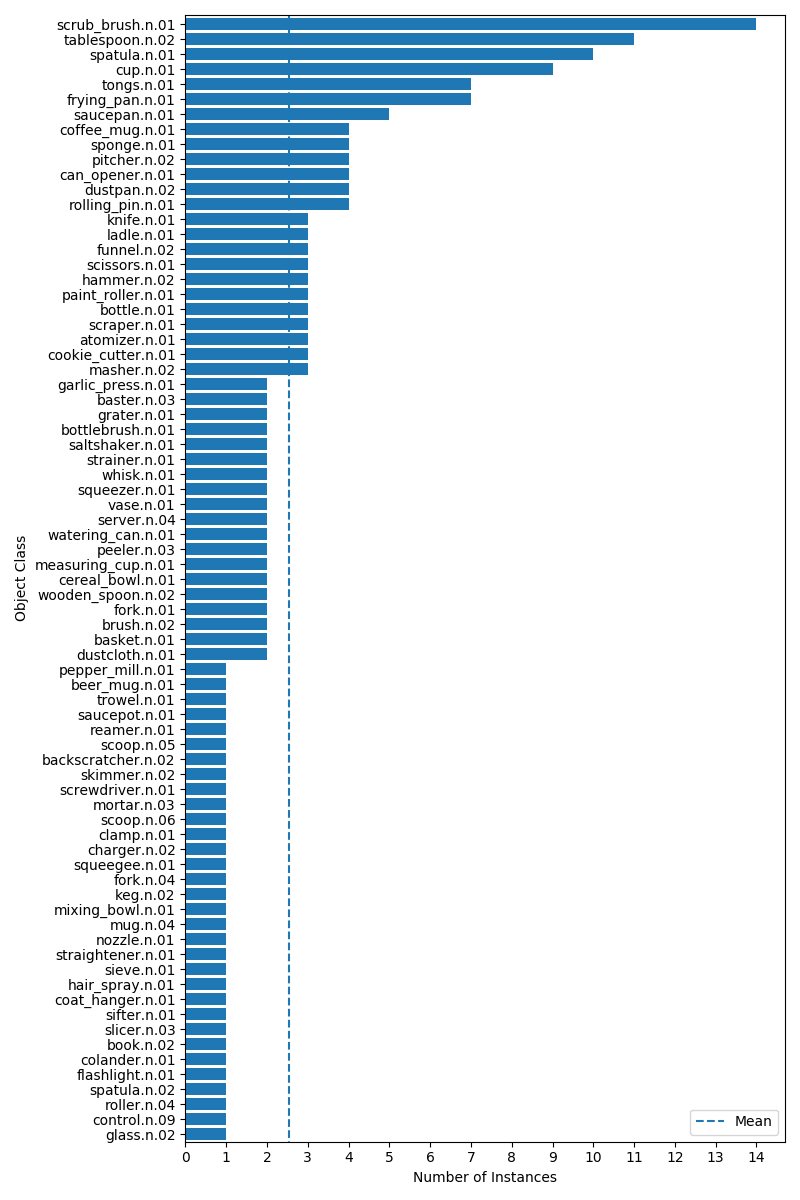}
  \caption{Number of object instances for each object category. The dotted vertical line is the average number of instances/category. The category is represented by its WordNet synset on the y-axis.
  \vspace{-4mm} }
\label{fig:histogram}
\end{figure}

\subsection{Additional Dataset Examples and Analysis}

Additional examples of task-oriented grasps for several objects are shown in Fig \ref{fig:extra_dataset}. Fig \ref{fig:histogram} shows a histogram of the number of object instances per object category. In \abvdataset, each object category is resolved to a WordNet synset \cite{miller1995wordnet} (displayed on the vertical axis). On average, each object category has approximately 2.5 instances. There are three main sets of categories based on the distribution: categories with high (cups, spatulas etc.), medium (paint roller, knife, etc.) and low (sifter, mixing bowl, etc) incidence. It is also noteworthy that our dataset is restricted to objects that can be effectively scanned by a depth sensor i.e. objects that are not transparent or too specular.

There are two main types of semantic knowledge provided in our dataset. The first type of knowledge are ``Used-For" relations between tasks and object categories. We list all the tasks suitable for each object category in Table \ref{tb:ot_pairs}. As explained in the paper, these task-object category combinations were collected in the first stage of annotation before labelling the task-oriented grasps. The second type of knowledge is the semantic hierarchy for objects, which includes a diverse set of object concepts interlinked by ``Is-A" relations. The hierarchy provides useful information for task-oriented grasping as objects in the same subtree of the hierarchy may share similar geometry or functionality. Table \ref{tb:object_paths} shows the WordNet hypernym paths for each object class, from the most specific concept (e.g. \textit{mug.n.01}) to the most abstract concept (e.g. \textit{entity.n.01}).

\subsection{Analysis on \abvmethod Predictions}

\begin{figure*}
\centering
    \includegraphics[width=0.65\linewidth]{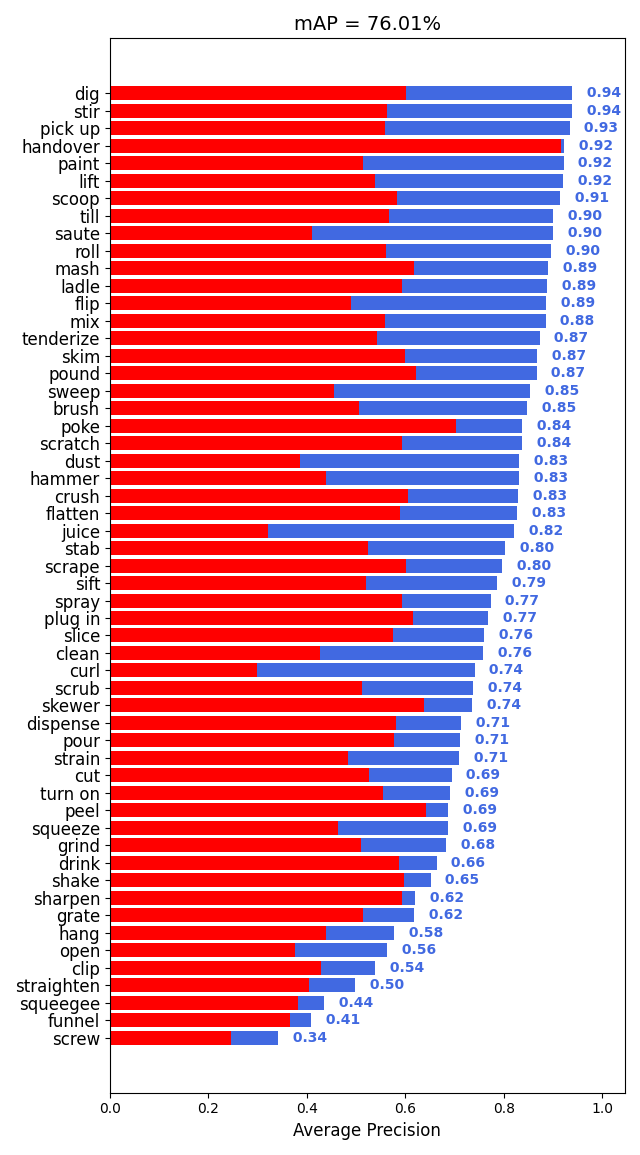}
  \caption{mAP across tasks for \abvmethod predictions. The red bar is for AP predictions by a random model while the red and blue cumulatively represents the model prediction.
  \vspace{-4mm} }
\label{fig:map_tasks}
\end{figure*}

Next we visualize AP scores for each task from \abvmethod predictions trained on \abvdataset in Fig \ref{fig:map_tasks}. The AP scores for all tasks were computed with cross validation as detailed in the main paper. The red bar corresponds to AP score with predictions from a random model (averaged over five seeds) while the red and blue bar cumulatively represents the model AP score. Overall, \abvmethod performed better than random predictions, though some tasks are more challenging than others. For instance, juice, saute and screw are harder tasks (with low random prediction scores) compared to handover and poke. Tasks that represent more creative than prototypical uses of an object are typically more ambiguous and challenging to label. Yet, our model is able to improve over random predictions even in these challenging tasks.



\begin{table}[t]
\footnotesize
\centering
\caption{Object Task Combinations}
\label{tb:ot_pairs}
\resizebox{\columnwidth}{!}{%
\begin{tabular}{ll}
\toprule
Object Class & Suitable Tasks \\
\midrule
atomizer.n.01 & pour, clean, squeegee, dispense, handover, spray \\
backscratcher.n.02 & pick up, turn on, shake, scrape, handover, scoop, scratch, mix \\
basket.n.01 & pick up, dispense, lift, handover, scoop, hang \\
baster.n.03 & poke, dispense, squeeze, handover, drink, spray \\
beer\_mug.n.01 & pour, scoop, ladle, flatten, dispense, hammer, lift, mash, handover, crush, drink, pound \\
book.n.02 & pound, crush, handover \\
bottle.n.01 & pour, clean, juice, squeegee, poke, shake, flip, flatten, dispense, hammer, straighten, handover, pound, drink, spray, mix \\
bottlebrush.n.01 & clean, dust, scrub, scrape, straighten, brush, mash, handover, crush, scratch \\
brush.n.02 & handover, brush, sweep, paint \\
can\_opener.n.01 & cut, open, squeegee, poke, pick up, squeeze, screw \\
cereal\_bowl.n.01 & pour, scoop, ladle, pick up, dispense, handover, drink, mix \\
charger.n.02 & plug in, turn on \\
clamp.n.01 & clip, straighten, lift, squeeze, crush, hang \\
coat\_hanger.n.01 & hang, straighten, pick up, handover \\
coffee\_mug.n.01 & pour, skim, clean, scoop, ladle, pick up, grind, shake, flatten, dispense, dig, lift, handover, drink, pound \\
colander.n.01 & pour, skim, juice, poke, dig, funnel, handover, crush, sift, scoop, strain, pound \\
control.n.09 & turn on \\
cookie\_cutter.n.01 & slice, cut \\
cup.n.01 & pour, skim, clean, scoop, ladle, till, saute, poke, pick up, shake, tenderize, flatten, dispense, dig, lift, crush, handover, pound, drink, mix \\
dustcloth.n.01 & clean, dust, brush, sweep \\
dustpan.n.02 & clean, dust, pick up, flip, dispense, lift, handover, scoop, sweep \\
flashlight.n.01 & turn on, handover \\
fork.n.01 & skewer, juice, ladle, poke, stir, pick up, handover, stab, flip, grate, dig, scrape, curl, lift, funnel, mash, scoop, scratch, strain, mix \\
fork.n.04 & till, stir, stab, dig, scrape, handover, scratch \\
frying\_pan.n.01 & pour, saute, stir, pick up, flip, tenderize, flatten, dispense, hammer, lift, mash, handover, crush, pound, scoop, mix \\
funnel.n.02 & pour, scoop, pick up, stab, dispense, scrape, squeeze, funnel, roll, strain, mix \\
garlic\_press.n.01 & grind, flatten, hammer, squeeze, mash, handover, crush, scratch \\
grater.n.01 & cut, slice, grind, tenderize, grate, scrape, scratch, strain \\
hair\_spray.n.01 & roll, spray, handover \\
hammer.n.02 & tenderize, flatten, hammer, straighten, mash, crush, pound \\
keg.n.02 & flatten, dispense, drink, pour \\
knife.n.01 & cut, peel, poke, slice, stab, clip, scrape, sharpen, scratch \\
ladle.n.01 & pour, skim, scoop, ladle, poke, saute, stir, pick up, dispense, hammer, scrape, lift, handover, sift, roll, drink, strain, sweep, mix \\
masher.n.02 & cut, juice, poke, stir, grind, tenderize, flatten, hammer, mash, handover, crush, pound, mix \\
measuring\_cup.n.01 & pour, scoop, ladle, pick up, dispense, dig, lift, handover, drink \\
mixing\_bowl.n.01 & pour, dispense, pick up, mix \\
mortar.n.03 & pour, stir, grind, tenderize, flatten, pound, mash, handover, crush, sift, scoop, mix \\
mug.n.04 & pour, drink, scoop, handover \\
nozzle.n.01 & dispense, spray \\
paint\_roller.n.01 & squeegee, paint, tenderize, flatten, dispense, brush, handover, roll, pound \\
pancake\_turner.n.01 & cut, skim, ladle, pick up, crush, scoop, saute, turn on, flatten, scrape, handover, mix, pound, pour, stir, poke, flip, dig, lift, mash, sift \\
peeler.n.03 & peel, slice, grate, scrape, scratch \\
pepper\_mill.n.01 & grind, crush, handover \\
pitcher.n.02 & pour, scoop, ladle, stir, pick up, shake, flatten, dispense, lift, handover, drink, mix \\
reamer.n.01 & plug in, juice, scrape, mash, handover \\
roller.n.04 & roll, clean, lift, handover \\
rolling\_pin.n.01 & poke, tenderize, flatten, hammer, straighten, mash, handover, crush, roll, pound \\
saltshaker.n.01 & pour, dispense, crush, handover, strain, shake \\
saucepan.n.01 & pour, scoop, ladle, saute, stir, pick up, shake, flatten, dispense, dig, lift, mash, handover, crush, drink, mix \\
saucepot.n.01 & pour, saute, dispense, lift, mash, handover, crush, drink, mix \\
scissors.n.01 & cut, open, poke, slice, handover, stab, clip, scrape, curl, straighten, sharpen, scratch \\
scoop.n.05 & clean, pick up, flip, dig, lift, handover, sift, scoop, strain, mix \\
scoop.n.06 & clean, stir, pick up, dig, lift, handover, scoop, pound \\
scraper.n.01 & peel, clean, squeegee, slice, stir, stab, flatten, dig, scrape, straighten, lift, handover, scoop, scratch \\
screwdriver.n.01 & skewer, open, poke, stab, dig, screw, hang, scratch, mix \\
scrub\_brush.n.01 & clean, dust, paint, poke, stir, scrub, shake, stab, flip, tenderize, flatten, scrape, straighten, mix, funnel, handover, brush, scratch, sweep, pound \\
server.n.04 & ladle, stir, pick up, curl, lift, handover, sift, scoop, hang, mix \\
sieve.n.01 & sift, dispense, strain, skim \\
sifter.n.01 & sift, dispense, strain \\
skimmer.n.02 & skim, ladle, saute, stir, pick up, flip, scrape, handover, sift, scoop, strain \\
slicer.n.03 & cut, peel, open, juice, slice, saute, grate, screw, mix \\
spatula.n.01 & skim, poke, saute, stir, pick up, scrub, flip, dig, lift, crush, handover, scoop, scratch, mix \\
spatula.n.02 & skim, stir, pick up, flip, flatten, scrape, lift, mix \\
sponge.n.01 & skim, clean, squeegee, dust, poke, scrub, scrape, brush, handover, drink, scratch, sweep \\
squeegee.n.01 & clean, squeegee, scrub, scrape, handover \\
squeezer.n.01 & juice, flatten, squeeze, mash, handover, crush, drink, pound \\
straightener.n.01 & flatten, pick up, straighten \\
strainer.n.01 & skim, stir, pick up, shake, flip, dispense, lift, funnel, handover, crush, sift, scoop, strain, sweep, mix \\
tablespoon.n.02 & skim, ladle, pick up, dispense, curl, scoop, scratch, saute, turn on, stab, flatten, scrape, squeeze, handover, mix, pound, pour, stir, poke, flip, dig, lift, mash, sift, drink, strain \\
tongs.n.01 & squeegee, pick up, clip, dispense, crush, scoop, scratch, saute, turn on, stab, scrape, squeeze, funnel, handover, shake, skewer, mix, stir, straighten, poke, flip, lift, roll \\
trowel.n.01 & till, slice, stir, poke, stab, flip, flatten, hammer, dig, scrape, lift, crush, scoop, scratch, mix \\
vase.n.01 & pour, scoop, tenderize, dig, straighten, lift, handover, drink, shake \\
watering\_can.n.01 & pour, scoop, poke, dispense, funnel, drink, shake \\
whisk.n.01 & mix, stir, brush, handover \\
wooden\_spoon.n.02 & skim, ladle, poke, saute, stir, pick up, flatten, dig, scrape, lift, mash, handover, pound, scoop, mix \\
\bottomrule
\end{tabular}}
\vspace{-.6cm}
\end{table}

\begin{table}[t]
\footnotesize
\centering
\caption{Object Hypernym Paths}
\label{tb:object_paths}
\resizebox{0.8\columnwidth}{!}{%
\begin{tabular}{lcl}
\toprule
Object Class & Number of Instances & Hypernym Paths \\
\midrule
atomizer.n.01 & 3 & atomizer.n.01$\rightarrow$container.n.01$\rightarrow$instrumentality.n.03$\rightarrow$entity.n.01 \\
backscratcher.n.02 & 1 & backscratcher.n.02$\rightarrow$device.n.01$\rightarrow$instrumentality.n.03$\rightarrow$entity.n.01 \\
basket.n.01 & 2 & basket.n.01$\rightarrow$container.n.01$\rightarrow$instrumentality.n.03$\rightarrow$entity.n.01 \\
baster.n.03 & 2 & baster.n.03$\rightarrow$utensil.n.01$\rightarrow$implement.n.01$\rightarrow$instrumentality.n.03$\rightarrow$entity.n.01 \\
beer\_mug.n.01 & 1 & beer\_mug.n.01$\rightarrow$mug.n.04$\rightarrow$vessel.n.03$\rightarrow$container.n.01$\rightarrow$instrumentality.n.03$\rightarrow$entity.n.01 \\
book.n.02 & 1 & book.n.02$\rightarrow$creation.n.02$\rightarrow$entity.n.01 \\
bottle.n.01 & 3 & bottle.n.01$\rightarrow$vessel.n.03$\rightarrow$container.n.01$\rightarrow$instrumentality.n.03$\rightarrow$entity.n.01 \\
bottlebrush.n.01 & 2 & bottlebrush.n.01$\rightarrow$brush.n.02$\rightarrow$implement.n.01$\rightarrow$instrumentality.n.03$\rightarrow$entity.n.01 \\
brush.n.02 & 2 & brush.n.02$\rightarrow$implement.n.01$\rightarrow$instrumentality.n.03$\rightarrow$entity.n.01 \\
can\_opener.n.01 & 4 & can\_opener.n.01$\rightarrow$tool.n.01$\rightarrow$implement.n.01$\rightarrow$instrumentality.n.03$\rightarrow$entity.n.01 \\
cereal\_bowl.n.01 & 2 & cereal\_bowl.n.01$\rightarrow$bowl.n.03$\rightarrow$container.n.01$\rightarrow$instrumentality.n.03$\rightarrow$entity.n.01 \\
cereal\_bowl.n.01 & 2 & cereal\_bowl.n.01$\rightarrow$bowl.n.03$\rightarrow$crockery.n.01$\rightarrow$ware.n.01$\rightarrow$article.n.02$\rightarrow$entity.n.01 \\
charger.n.02 & 1 & charger.n.02$\rightarrow$device.n.01$\rightarrow$instrumentality.n.03$\rightarrow$entity.n.01 \\
clamp.n.01 & 1 & clamp.n.01$\rightarrow$device.n.01$\rightarrow$instrumentality.n.03$\rightarrow$entity.n.01 \\
coat\_hanger.n.01 & 1 & coat\_hanger.n.01$\rightarrow$device.n.01$\rightarrow$instrumentality.n.03$\rightarrow$entity.n.01 \\
coffee\_mug.n.01 & 4 & coffee\_mug.n.01$\rightarrow$mug.n.04$\rightarrow$vessel.n.03$\rightarrow$container.n.01$\rightarrow$instrumentality.n.03$\rightarrow$entity.n.01 \\
colander.n.01 & 2 & colander.n.01$\rightarrow$strainer.n.01$\rightarrow$device.n.01$\rightarrow$instrumentality.n.03$\rightarrow$entity.n.01 \\
control.n.09 & 1 & control.n.09$\rightarrow$device.n.01$\rightarrow$instrumentality.n.03$\rightarrow$entity.n.01 \\
cookie\_cutter.n.01 & 3 & cookie\_cutter.n.01$\rightarrow$utensil.n.01$\rightarrow$implement.n.01$\rightarrow$instrumentality.n.03$\rightarrow$entity.n.01 \\
cup.n.01 & 10 & cup.n.01$\rightarrow$container.n.01$\rightarrow$instrumentality.n.03$\rightarrow$entity.n.01 \\
cup.n.01 & 10 & cup.n.01$\rightarrow$crockery.n.01$\rightarrow$ware.n.01$\rightarrow$article.n.02$\rightarrow$entity.n.01 \\
dustcloth.n.01 & 2 & dustcloth.n.01$\rightarrow$fabric.n.01$\rightarrow$entity.n.01 \\
dustcloth.n.01 & 2 & dustcloth.n.01$\rightarrow$part.n.02$\rightarrow$entity.n.01 \\
dustpan.n.02 & 4 & dustpan.n.02$\rightarrow$container.n.01$\rightarrow$instrumentality.n.03$\rightarrow$entity.n.01 \\
flashlight.n.01 & 1 & flashlight.n.01$\rightarrow$device.n.01$\rightarrow$instrumentality.n.03$\rightarrow$entity.n.01 \\
fork.n.01 & 2 & fork.n.01$\rightarrow$cutlery.n.02$\rightarrow$ware.n.01$\rightarrow$article.n.02$\rightarrow$entity.n.01 \\
fork.n.04 & 1 & fork.n.04$\rightarrow$tool.n.01$\rightarrow$implement.n.01$\rightarrow$instrumentality.n.03$\rightarrow$entity.n.01 \\
frying\_pan.n.01 & 7 & frying\_pan.n.01$\rightarrow$pan.n.01$\rightarrow$utensil.n.01$\rightarrow$implement.n.01$\rightarrow$instrumentality.n.03$\rightarrow$entity.n.01 \\
funnel.n.02 & 3 & funnel.n.02$\rightarrow$utensil.n.01$\rightarrow$implement.n.01$\rightarrow$instrumentality.n.03$\rightarrow$entity.n.01 \\
garlic\_press.n.01 & 2 & garlic\_press.n.01$\rightarrow$machine.n.01$\rightarrow$device.n.01$\rightarrow$instrumentality.n.03$\rightarrow$entity.n.01 \\
grater.n.01 & 2 & grater.n.01$\rightarrow$utensil.n.01$\rightarrow$implement.n.01$\rightarrow$instrumentality.n.03$\rightarrow$entity.n.01 \\
hair\_spray.n.01 & 1 & hair\_spray.n.01$\rightarrow$instrumentality.n.03$\rightarrow$entity.n.01 \\
hammer.n.02 & 3 & hammer.n.02$\rightarrow$tool.n.01$\rightarrow$implement.n.01$\rightarrow$instrumentality.n.03$\rightarrow$entity.n.01 \\
keg.n.02 & 1 & keg.n.02$\rightarrow$vessel.n.03$\rightarrow$container.n.01$\rightarrow$instrumentality.n.03$\rightarrow$entity.n.01 \\
knife.n.01 & 3 & knife.n.01$\rightarrow$tool.n.01$\rightarrow$implement.n.01$\rightarrow$instrumentality.n.03$\rightarrow$entity.n.01 \\
ladle.n.01 & 3 & ladle.n.01$\rightarrow$vessel.n.03$\rightarrow$container.n.01$\rightarrow$instrumentality.n.03$\rightarrow$entity.n.01 \\
masher.n.02 & 3 & masher.n.02$\rightarrow$utensil.n.01$\rightarrow$implement.n.01$\rightarrow$instrumentality.n.03$\rightarrow$entity.n.01 \\
measuring\_cup.n.01 & 2 & measuring\_cup.n.01$\rightarrow$container.n.01$\rightarrow$instrumentality.n.03$\rightarrow$entity.n.01 \\
mixing\_bowl.n.01 & 1 & mixing\_bowl.n.01$\rightarrow$bowl.n.03$\rightarrow$container.n.01$\rightarrow$instrumentality.n.03$\rightarrow$entity.n.01 \\
mixing\_bowl.n.01 & 1 & mixing\_bowl.n.01$\rightarrow$bowl.n.03$\rightarrow$crockery.n.01$\rightarrow$ware.n.01$\rightarrow$article.n.02$\rightarrow$entity.n.01 \\
mortar.n.03 & 1 & mortar.n.03$\rightarrow$vessel.n.03$\rightarrow$container.n.01$\rightarrow$instrumentality.n.03$\rightarrow$entity.n.01 \\
mug.n.04 & 1 & mug.n.04$\rightarrow$vessel.n.03$\rightarrow$container.n.01$\rightarrow$instrumentality.n.03$\rightarrow$entity.n.01 \\
nozzle.n.01 & 1 & nozzle.n.01$\rightarrow$opening.n.10$\rightarrow$entity.n.01 \\
paint\_roller.n.01 & 3 & paint\_roller.n.01$\rightarrow$roller.n.04$\rightarrow$shape.n.02$\rightarrow$entity.n.01 \\
pancake\_turner.n.01 & 8 & pancake\_turner.n.01$\rightarrow$turner.n.08$\rightarrow$utensil.n.01$\rightarrow$implement.n.01$\rightarrow$instrumentality.n.03$\rightarrow$entity.n.01 \\
peeler.n.03 & 2 & peeler.n.03$\rightarrow$device.n.01$\rightarrow$instrumentality.n.03$\rightarrow$entity.n.01 \\
pepper\_mill.n.01 & 1 & pepper\_mill.n.01$\rightarrow$machine.n.01$\rightarrow$device.n.01$\rightarrow$instrumentality.n.03$\rightarrow$entity.n.01 \\
pitcher.n.02 & 4 & pitcher.n.02$\rightarrow$vessel.n.03$\rightarrow$container.n.01$\rightarrow$instrumentality.n.03$\rightarrow$entity.n.01 \\
reamer.n.01 & 1 & reamer.n.01$\rightarrow$squeezer.n.01$\rightarrow$utensil.n.01$\rightarrow$implement.n.01$\rightarrow$instrumentality.n.03$\rightarrow$entity.n.01 \\
roller.n.04 & 1 & roller.n.04$\rightarrow$shape.n.02$\rightarrow$entity.n.01 \\
rolling\_pin.n.01 & 4 & rolling\_pin.n.01$\rightarrow$utensil.n.01$\rightarrow$implement.n.01$\rightarrow$instrumentality.n.03$\rightarrow$entity.n.01 \\
saltshaker.n.01 & 2 & saltshaker.n.01$\rightarrow$container.n.01$\rightarrow$instrumentality.n.03$\rightarrow$entity.n.01 \\
saucepan.n.01 & 5 & saucepan.n.01$\rightarrow$pan.n.01$\rightarrow$utensil.n.01$\rightarrow$implement.n.01$\rightarrow$instrumentality.n.03$\rightarrow$entity.n.01 \\
saucepot.n.01 & 1 & saucepot.n.01$\rightarrow$utensil.n.01$\rightarrow$implement.n.01$\rightarrow$instrumentality.n.03$\rightarrow$entity.n.01 \\
saucepot.n.01 & 1 & saucepot.n.01$\rightarrow$vessel.n.03$\rightarrow$container.n.01$\rightarrow$instrumentality.n.03$\rightarrow$entity.n.01 \\
scissors.n.01 & 3 & scissors.n.01$\rightarrow$implement.n.01$\rightarrow$instrumentality.n.03$\rightarrow$entity.n.01 \\
scissors.n.01 & 3 & scissors.n.01$\rightarrow$tool.n.01$\rightarrow$implement.n.01$\rightarrow$instrumentality.n.03$\rightarrow$entity.n.01 \\
scoop.n.05 & 1 & scoop.n.05$\rightarrow$tool.n.01$\rightarrow$implement.n.01$\rightarrow$instrumentality.n.03$\rightarrow$entity.n.01 \\
scoop.n.06 & 1 & scoop.n.06$\rightarrow$ladle.n.01$\rightarrow$vessel.n.03$\rightarrow$container.n.01$\rightarrow$instrumentality.n.03$\rightarrow$entity.n.01 \\
scraper.n.01 & 3 & scraper.n.01$\rightarrow$tool.n.01$\rightarrow$implement.n.01$\rightarrow$instrumentality.n.03$\rightarrow$entity.n.01 \\
screwdriver.n.01 & 1 & screwdriver.n.01$\rightarrow$tool.n.01$\rightarrow$implement.n.01$\rightarrow$instrumentality.n.03$\rightarrow$entity.n.01 \\
scrub\_brush.n.01 & 14 & scrub\_brush.n.01$\rightarrow$brush.n.02$\rightarrow$implement.n.01$\rightarrow$instrumentality.n.03$\rightarrow$entity.n.01 \\
server.n.04 & 2 & server.n.04$\rightarrow$utensil.n.01$\rightarrow$implement.n.01$\rightarrow$instrumentality.n.03$\rightarrow$entity.n.01 \\
sieve.n.01 & 1 & sieve.n.01$\rightarrow$strainer.n.01$\rightarrow$device.n.01$\rightarrow$instrumentality.n.03$\rightarrow$entity.n.01 \\
sifter.n.01 & 1 & sifter.n.01$\rightarrow$sieve.n.01$\rightarrow$strainer.n.01$\rightarrow$device.n.01$\rightarrow$instrumentality.n.03$\rightarrow$entity.n.01 \\
skimmer.n.02 & 1 & skimmer.n.02$\rightarrow$utensil.n.01$\rightarrow$implement.n.01$\rightarrow$instrumentality.n.03$\rightarrow$entity.n.01 \\
slicer.n.03 & 1 & slicer.n.03$\rightarrow$knife.n.01$\rightarrow$tool.n.01$\rightarrow$implement.n.01$\rightarrow$instrumentality.n.03$\rightarrow$entity.n.01 \\
spatula.n.01 & 2 & spatula.n.01$\rightarrow$turner.n.08$\rightarrow$utensil.n.01$\rightarrow$implement.n.01$\rightarrow$instrumentality.n.03$\rightarrow$entity.n.01 \\
spatula.n.02 & 1 & spatula.n.02$\rightarrow$tool.n.01$\rightarrow$implement.n.01$\rightarrow$instrumentality.n.03$\rightarrow$entity.n.01 \\
sponge.n.01 & 4 & sponge.n.01$\rightarrow$matter.n.03$\rightarrow$entity.n.01 \\
sponge.n.01 & 4 & sponge.n.01$\rightarrow$relation.n.01$\rightarrow$entity.n.01 \\
squeegee.n.01 & 1 & squeegee.n.01$\rightarrow$implement.n.01$\rightarrow$instrumentality.n.03$\rightarrow$entity.n.01 \\
squeezer.n.01 & 2 & squeezer.n.01$\rightarrow$utensil.n.01$\rightarrow$implement.n.01$\rightarrow$instrumentality.n.03$\rightarrow$entity.n.01 \\
straightener.n.01 & 1 & straightener.n.01$\rightarrow$device.n.01$\rightarrow$instrumentality.n.03$\rightarrow$entity.n.01 \\
strainer.n.01 & 1 & strainer.n.01$\rightarrow$device.n.01$\rightarrow$instrumentality.n.03$\rightarrow$entity.n.01 \\
tablespoon.n.02 & 11 & tablespoon.n.02$\rightarrow$spoon.n.01$\rightarrow$cutlery.n.02$\rightarrow$ware.n.01$\rightarrow$article.n.02$\rightarrow$entity.n.01 \\
tablespoon.n.02 & 11 & tablespoon.n.02$\rightarrow$spoon.n.01$\rightarrow$container.n.01$\rightarrow$instrumentality.n.03$\rightarrow$entity.n.01 \\
tongs.n.01 & 7 & tongs.n.01$\rightarrow$device.n.01$\rightarrow$instrumentality.n.03$\rightarrow$entity.n.01 \\
trowel.n.01 & 1 & trowel.n.01$\rightarrow$tool.n.01$\rightarrow$implement.n.01$\rightarrow$instrumentality.n.03$\rightarrow$entity.n.01 \\
vase.n.01 & 2 & vase.n.01$\rightarrow$vessel.n.03$\rightarrow$container.n.01$\rightarrow$instrumentality.n.03$\rightarrow$entity.n.01 \\
watering\_can.n.01 & 2 & watering\_can.n.01$\rightarrow$container.n.01$\rightarrow$instrumentality.n.03$\rightarrow$entity.n.01 \\
whisk.n.01 & 2 & whisk.n.01$\rightarrow$utensil.n.01$\rightarrow$implement.n.01$\rightarrow$instrumentality.n.03$\rightarrow$entity.n.01 \\
wooden\_spoon.n.02 & 2 & wooden\_spoon.n.02$\rightarrow$spoon.n.01$\rightarrow$cutlery.n.02$\rightarrow$ware.n.01$\rightarrow$article.n.02$\rightarrow$entity.n.01 \\
wooden\_spoon.n.02 & 2 & wooden\_spoon.n.02$\rightarrow$spoon.n.01$\rightarrow$container.n.01$\rightarrow$instrumentality.n.03$\rightarrow$entity.n.01 \\
wooden\_spoon.n.02 & 2 & wooden\_spoon.n.02$\rightarrow$ware.n.01$\rightarrow$article.n.02$\rightarrow$entity.n.01 \\
\bottomrule
\end{tabular}}
\vspace{-.6cm}
\end{table}

\subsection{Annotation Interface on Amazon Mechanical Turk}
We now describe the process for annotating task-oriented grasps for a given object and task. Instead of manually labelling the dataset like prior work \cite{liuICRA2020}, we scale the labelling effort for our larger dataset using Amazon Mechanical Turk (AMT). As explained in the main paper, we have a two-stage procedure to reduce the annotation cost.

\begin{figure*}
\centering
    \includegraphics[width=0.65\linewidth]{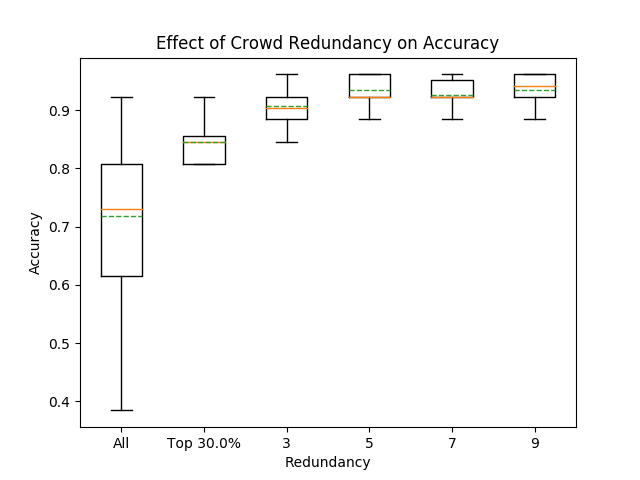}
  \caption{Results of qualification test. The left-most entry is the results for all participants in the test. The box for each entry extends from the lower to upper quartile value, with the dotted and solid line at the mean and median accuracy respectively. The second entry from the left is the top 30\% of the crowd workers who are recruited for the actual annotation. For the remaining entries, 10 sets of crowd turkers were randomly selected from the top 30\% cohort with a redundant set size of K =3,5,7,9. The average accuracy for each set size is plotted with the performance saturating after K=3. As such, we used three annotators for labelling each task-oriented grasp.}
\label{fig:qualification}
\end{figure*}

\begin{figure*}
\centering
    \includegraphics[width=0.85\linewidth]{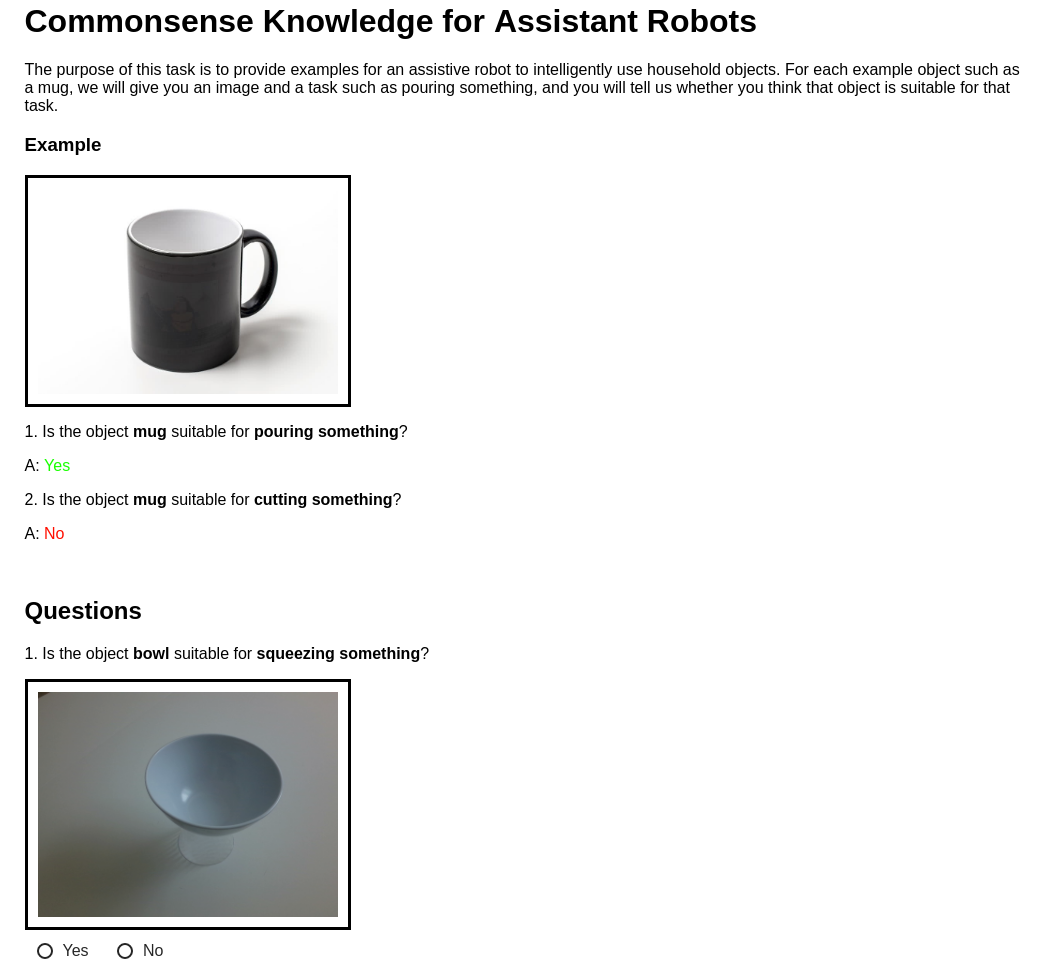}
  \caption{Example from the first annotation stage to gather labels for task suitability for each object.
  \vspace{-4mm} }
\label{fig:mturk_stage1}
\end{figure*}

In the first stage, we gathered suitable tasks for each object.  In each of our labeling tasks, we presented annotators with the image of the object, the object category, and the task, as shown in the example in Fig \ref{fig:mturk_stage1}.

\begin{figure*}
\centering
    \includegraphics[width=0.85\linewidth]{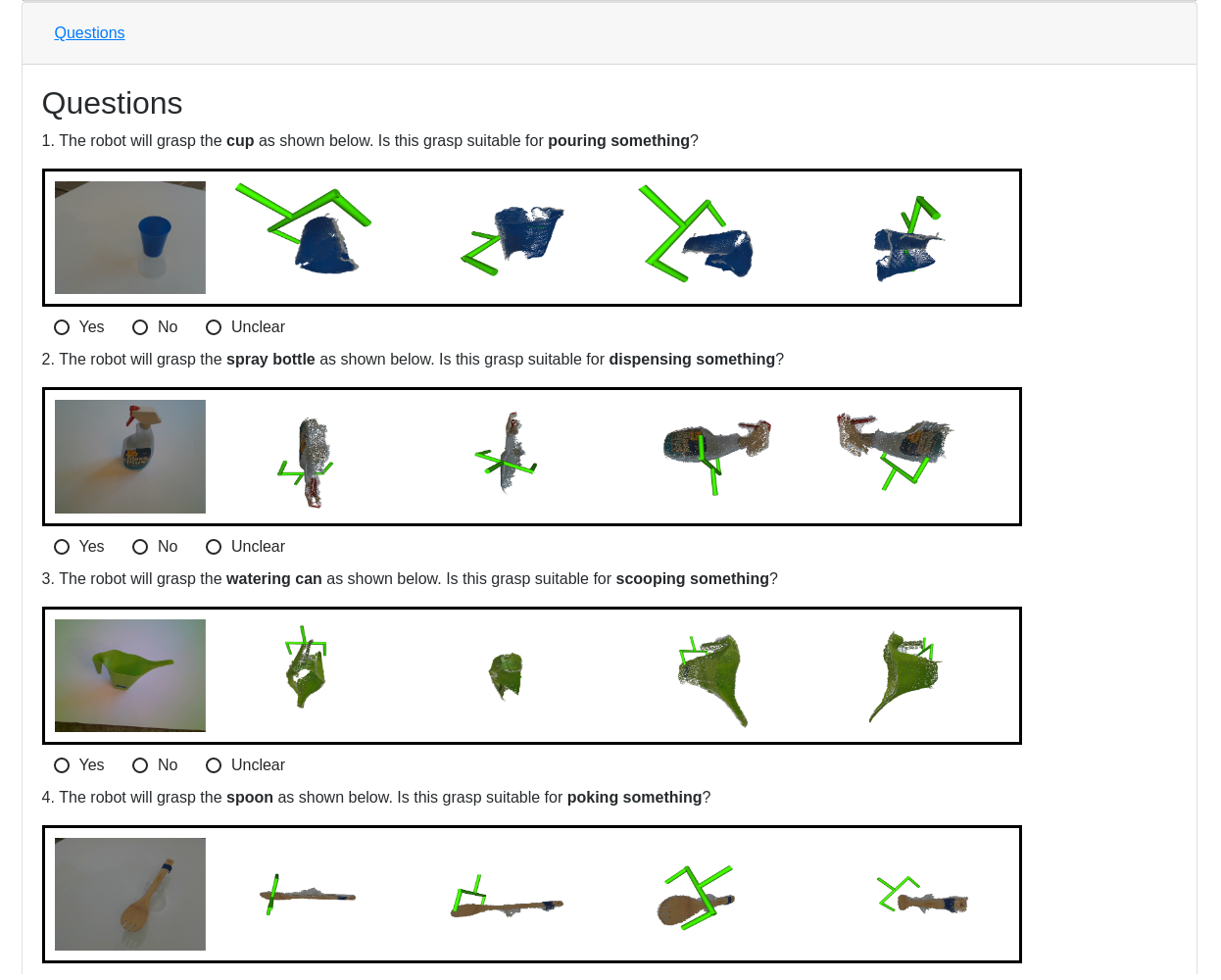}
  \caption{Example questions from the second stage of annotation to label task-oriented grasps.
  \vspace{-4mm} }
\label{fig:mturk_stage2}
\end{figure*}

In the second stage, we collected labels for task-oriented grasping only if the task applies to the object (filtered from the first stage). For each grasp, we presented the annotators with an image of the object, visualizations of the object point cloud and the grasp from 3 different angles, the object category, and the task. Examples of this graphical interface are shown in Fig \ref{fig:mturk_stage2}. Since labelling on 3D data from a 2D interface is highly ambiguous and challenging, we presented visualizations from multiple views.

In our pilot study, we found that annotators do not always agree with each other since the notion of task-oriented grasping for robots is generally ambiguous to non-expert users.  As a result, we provided them with seven guidelines concerning different aspects of task-oriented grasps, such as functionality, stability, creative uses of objects and safety. We empirically found that most crowd workers were able to understand the annotation task, though it takes some time to internalize the object shape and task description before deciding on the label. We used a qualification test to recruit crowd workers. The test had 26 questions which the authors annotated the ground-truths for. Fig \ref{fig:qualification} shows the results from the qualification test. We rank the test participants and qualify the top 30\% for the final annotation round. To improve annotation quality, each task-oriented grasp is presented to three annotators and the final label is decided on a majority vote. As shown in Fig \ref{fig:qualification}, we found that having a redundant set of three crowd workers for each question was a good trade-off between annotation cost (which scales linearly with redundancy) and label quality. Both the guidelines and the qualification test effectively improved the quality and consistency of labels.

\end{document}